# Meaning and understanding in large language models

Vladimír Havlík
Institute of Philosophy of the Czech Academy of Sciences, Prague
University of West Bohemia, Pilsen

## Abstract

Can a machine understand the meanings of natural language? Recent developments in the generative large language models (LLMs) of artificial intelligence have led to the belief that traditional philosophical assumptions about machine understanding of language need to be revised. This article critically evaluates the prevailing tendency to regard machine language performance as mere *syntactic manipulation* and the *simulation* of understanding, which is only *partial* and very *shallow*, without sufficient *referential grounding* in the world. The aim is to highlight the conditions crucial to attributing natural language understanding to state-of-the-art LLMs, where it can be legitimately argued that LLMs not only use syntax but also *semantics*, their understanding not being simulated but *duplicated*; and determine how they *ground* the meanings of linguistic expressions.

## Introduction

A notable feature of the current generative AI boom is the machine processing of natural language by LLMs. The best language model architectures—transformers—work with natural language indistinguishable from that of humans in many different language activities: translation, text generation and summarization, meaning and emotional colour matching, dialogue, and other "language games." They thus possess a capacity for natural language that has hitherto been associated exclusively with humans. A controversial issue, given the successes of natural language processing (NLP) machines, is the question of their understanding of natural language. Can a machine understand the meanings of the language through which machine-human communication takes place? A state-of-the-art generative AI model leads to the belief that traditional philosophical assumptions about language understanding need to be revised. This presupposes a critical evaluation of the prevailing tendency to regard machine language performance as mere *syntactic manipulation* and the *simulation of* understanding, which is only *partial* and very *shallow*, without sufficient *referential grounding* in the world. The aim of this article is to highlight the conditions crucial to attributing natural language understanding to LLMs, where it can be legitimately argued that LLMs not only use syntax but also *semantics*, their understanding not being simulated but *duplicated*; and determine how they *ground* the meanings of linguistic expressions.

The paper is organized into three basic parts, as follows: in the first part I address the problem of language understanding and the validity of some traditional assumptions about the biological determination of brain properties, including language understanding. The current functionality of artificial neural networks (ANNs) in recognition and generation capabilities is then taken as a contemporary form of empirical evidence for the invalidity of some such beliefs. In the second part, I address the problem of the relationship between syntax and semantics and attempt to demonstrate their interdependence by formulating a thesis on *minimal semantic contents*. Then, in the third and final part, the relation between syntax and semantics turns into "the symbol grounding problem" and its modern variation, "the vector grounding problem" within state-of-the-art LLMs of generative AI. I conclude by showing how meanings are grounded in the LLMs and how all the conditions discussed are met, which entitle us to attribute natural language understanding not only to humans but also to machines.

# 1 Understanding language

The mastery of natural language is a specific process that requires not only genetic prerequisites but also cultural and social conditions and, as a specifically developed ability, is attributed only to humans. It is therefore surprising that we have been able to transfer this ability to a non-organic carrier and gradually reach a level where language processing in LLMs by generative AI is indistinguishable from that by humans. If the results were not compelling and unique, we would hardly dare to claim that the systems actually understand language. However, the persuasiveness and uniqueness of the generated responses alone are not enough to claim that systems *understand* our language. What would be meant by such "understanding"?

## 1.1 Simulation and replication of understanding

If, for example, I claim that systems *understand* textual assignments like "red circle on the sea" because they are able to generate images that my sensory-linguistic experience evaluates as "red circles on the sea", each of the responses being unique because they preserve the *sense* of the proposition but *vary* the redness, circularity, and sea *in a unique* way, then there is always the possibility of claiming that the system does not really understand anything but merely *simulates* understanding well. In other words, it behaves *as if it* understands, but in fact does not operate at all on the meanings of particular concepts such as redness, circularity, or sea. The resulting image is generated according to pre-set, previously learned stochastic patterns stored in "weights" at each level and node of the neural network, learned from many text-image data. Such a system behaves in a similar way in the case of text generation. One putative explanation, then, is that however the language models generate meaningful text, the one who gives it meaning is a human, because the system merely assembles words into sentences based on learned probabilities, and thus, according to some interpretations, the system behaves like a kind of "stochastic parrot" (Bender et al. 2021).

John Searle took a similar position in the early 1980s with respect to the possibility of strong AI, including the claim that "no simulation by itself ever constitutes duplication" (Searle 1984). This approach is currently maintained by some linguists, philosophers and computer scientists such as Bender et al. (2021), Langrebe and Smith (2019), Browning and Lecun (2022), Hadfield (2022) and others who do not see the performance of LLMs as actually understanding linguistic meanings. In these cases, the authors are inclined to the view that these systems of understanding are more likely to mimic (i.e. a *kind of mimicry*) because their understanding of language, while it may be impressive, is *shallow*. They then compare the performance of LLMs to, for example, the jargon-spouting students who try to imitate their professors but basically do not know what they are talking about (Browning and Lecun 2022).

A significantly different position on the hypothetical spectrum of language understanding is advocated by Anders Søgaard (2022) who subscribes to Dennett's (1987) critique of Searle, and attempts to show that consciousness is not a necessary condition for language understanding, given the "empirical observation that language understanding can be unconscious" (Søgaard 2022). Language models are therefore able to learn partially natural language, and Søgaard tries to show that under certain conditions not only inferential but also referential semantics is available to them in this sense. This does not mean, however, that language models are able to make use of language in the same variety of ways as humans, who are embedded in the world and benefit from causal contexts synchronized with their sensory experience. Søgaard therefore suggests that even though language models play many language games better than we humans do, ostensive definitions, or the *pointing game*, remain a crucial factor in understanding language. Thus, in a sense, LLMs *understand* natural language because understanding is not contingent on an *awareness* of that understanding. Understanding is not simulated but duplicated in the generated textual propositions, and although referential semantics is theoretically possible, achieving it is more challenging for LLMs.

Towards the other end of the language understanding spectrum, efforts have been made to show that the LLMs' language understanding is already so convincing that the generated responses of sufficiently trained AI systems demonstrate the convincing autonomy of a sentient individual who exhibits a degree of awareness and self-awareness by emphasizing in their responses their emotions and feelings in experiencing their existence (Lemoine 2022).[1] Does this mean that such systems not only understand natural language but also exhibit some form of phenomenal consciousness? That is, that they are carriers of their own mental states, beliefs and intentions and that they are capable of reflecting them? In such a case, understanding would not only be *duplicated* because the system actually *understands the* propositions given, but would also have a causal effect on the continuously maintained representation of the development of the dialogue, i.e. what we might call an impression, a feeling, a kind of emotional trace that the dialogue leaves in its bearer.

The distinction between *simulation* and *duplication* is thus a crucial criterion for understanding language models that are able to participate with us in language games. Unfortunately, however, there is no unambiguous criterion that can determine independently of us whether a system only behaves as if it understands or whether it actually understands. If we are unable to decide between understanding and "as if" understanding, then we are also unable to decide whether the system is merely simulating the language game or actually duplicating it. It seems, then, that we need some other empirical criterion whose presence or absence would allow us to decide whether it is just a simulation or a duplication. Let us look at this dilemma from another angle.

If we say that current AI systems *understand*, or more carefully, are capable of some degree of *understanding* of natural human language, what does that *understanding* actually mean? Is it to be relegated to the activation/deactivation of pre-given rules, or is this implementation of rules transcended by some internal autonomous representations that are to some extent independent of the elementary states of the machine, much as our thoughts are to some extent independent of the state of a particular neuron? Initially, we must start with the assumption that the problem may lie in our inability to define the term *understand* well enough. So let us ask, *how do we understand that something, or someone, understands something?*

## 1.2 Understanding "understanding"

For example, would we say that an AI system understands natural language if we ask it to generate an image based on the text input "red circle on the sea", and the system has the following options: 1) colours: red, yellow, green; 2) geometric shapes: circle, square, triangle; 3) background: sea, meadow, forest? If the system correctly chooses the colour, the geometric shape and the desired background, in this case 1a,2a,3a, does it mean that it *understands* our language? Intuition tells us that it does not, because the shapes, colours and background are given and only allow for other possible combinations of each other 1b,2a,3a, ..., 1c,2c,3c. Understanding the *meanings* of the individual terms plays no decisive role here. A simple comparison of the input values with the values that are available in advance is sufficient to generate the image. The system thus has no intention of the colour red, or any other colour, let alone a moving circle on the sea. The image is just one of the combinations available to the system, and its generation is the execution of an algorithm that must be determined in advance. Our intuition thus rightly counters the claim that the system responds based on its *understanding* of language.

---

[1] A well-publicized event from Google's mid-2022.

This elementary example shows how we would be forced to think about the possible understanding of a system within a symbolic approach to AI. The fundamental difference in approaches between symbolic AI and neural network models lies in the extent to which we try to determine the process required to obtain the desired behavioural outcome. "Put simply, the logic-inspired paradigm views sequential reasoning as the essence of intelligence and aims to implement reasoning in computers using hand-designed rules of inference that operate on hand-designed symbolic expressions that formalize knowledge. The brain-inspired paradigm views learning representations from data as the essence of intelligence and aims to implement learning by hand-designing or evolving rules for modifying the connection strengths in simulated networks of artificial neurons." (Bengio, Lecun, and Hinton 2021)

Fundamental differences in these approaches are also evident in the field of natural language processing. Although the symbolic approach is capable of algorithmizing many NLP processes, such as extracting key information, classifying, labelling, and structuring previously unstructured data, it does not allow for the process of learning from data, and thus the implementation of complex, hard-to-algorithmize tasks is difficult. In contrast, language models of various architectures using the concept of neural networks and the process of *deep learning* are able to learn and retain the recognized regularities and patterns in the language and follow them in other cases. However, the obvious difference between symbolic and neural AI approaches does not yet tell us anything about understanding language. If systems using neural networks have any advantage in NLP/NLU, then these contexts need to be justified in more detail and possibly arguments against the possibility of language understanding need to be refuted.

So what is going on differently and specifically in our organic brain when we generate an image based on the "red circle on the sea" task, as opposed to the inorganic language model of the neural network learned in the *deep learning* process? This is obviously a fundamental question, but not one that is being asked for the first time. If we want to try to answer it at the current state of AI development, we must first go back to the original argument.

## 1.3 The Chinese room argument

The most influential attempt to counter the possibility of strong AI and the other accompanying specific brain properties, including understanding, was formulated by John Searle in the 1980s as the *Chinese room argument* (Searle 1980). Although the wave of controversy the argument has generated seems to have peaked, most of the questions remain open. At that time, practical attempts at AI were mainly focused on the development of symbolic AI, as there was not yet the appropriate technology or enough suitable data for a *neural* connectionist approach. At the time, proponents[2] of strong AI argued that intelligence was nothing more than "the ability to manipulate symbols" (Searle 1984) and need not be tied exclusively to the brain and its biological origins, but as an algorithmizable process, it could also be achieved on a non-biological basis, e.g., in a digital computer. Searle, on the other hand, was convinced that he could present a decisive argument to show that while "symbol manipulation" may be successful and convincing, the system may not be intelligent. He therefore devised a simple thought experiment in which he, as a person unfamiliar with Chinese, is confined to a room (the system) and takes in Chinese characters (the input), and according to a detailed manual (the program) in a language he understands, arranges other Chinese characters (the symbol manipulation) and feeds them out (the output). The system as a whole manipulates the Chinese characters as if it understands them, e.g. it responds to a given question in Chinese by constructing convincing answers that a native Chinese speaker would not recognize as being answered by someone who does not understand Chinese at all. The system, including Searle as the central unit, would thus only *simulate* understanding Chinese, but not *duplicate* it with its own understanding.

---

[2] E.g. Herbert Simon, Alan Newell, Marvin Minsky, John McCarthy.

Thus, we see that the issue of simulation and duplication is again relevant in the context of current AI performance, and Searle's belief that computer systems can only achieve simulation is still the most acceptable explanation for many. Although Searle's argument has been subjected to extensive criticism from various theoretical positions (see Searle 1990), the existence of generative AI systems now makes it possible to empirically reject some of Searle's assumptions. Although it might seem that the different technological context of the 1980s did not allow Searle to sufficiently account for the future functionality of neural networks[3] (e.g., the performance of current convolutional neural networks in computer image recognition or transformational language models in language understanding), he formulated his argument in a way that was independent of technological developments, and I therefore consider it important to highlight this aspect in more detail.

For example, when evaluating the abilities of the brain and the computer, Searle argued that a computer would have to perform an enormous amount of computation on geometric and topological features to recognize a face, whereas we only need a glance to recognize it. Moreover, we have no evidence that our brains do any computation at all in the case of facial recognition (Searle 1984 Chap. III.). Searle therefore considered certain properties of the brain to be specifically biological and thought that they could either not be achieved at all on a non-biological basis or only with great difficulty. *Recognition* would be burdened with computations that the brain does not perform at all, *understanding* could not be true understanding because the computer has only syntax and not semantics, *intentionality* cannot occur within any program, and like *thinking,* all of these abilities are biologically determined and as such are the result of *internal causal forces* of the *brain* (Searle 1980). As I have already pointed out, although Searle could not assume the capabilities of neural networks, he still dared to claim that his argument was not dependent on *technological developments* if we proceed from the definition of *a digital* computer (Searle 1984).

His argument now faces two important facts. 1) Neural network models now perform recognition (and many other tasks associated with computer vision) in a way very similar to the organic brain, without a lot of calculating of geometrical and topological features. They creatively generate texts and images indistinguishable those of humans on the basis of textual or graphical inputs, and it is thus evident that these properties belong to the brain not because it is a specifically biological organ, but because we have used its structurally similar arrangement (the neural network model) and the corresponding deep learning functionality to implement them. 2) However, neural network models do not necessarily require a *hardware* structure derived from the neural networks of the brain, but the structure of the interconnected layers of neurons is achievable virtually in a classical *digital* computer (i.e., von Neumann architecture), i.e., in simplified terms, by a *program*. The important conclusion, then, is that not all brain functionalities are specifically biological, as Searle assumed, and can be achieved, contrary to Searle's assumptions, *by implementing a suitable program*. The existence of neural networks and their properties can be taken as *empirical evidence* of the invalidity of Searle's argument against the possibility of strong artificial intelligence. Thus, this empirical fact weakens the belief of biological naturalism about the fundamental biological nature of the aforementioned brain abilities and testifies in favour of connectionism based on the specifically structural arrangement of neural networks.

## 2   The gap between syntax and semantics

The point of Searle's argument was that "The computer has syntax but not semantics", a claim he repeats in many places (e.g. Searle 1980, 423) whereas our organic brains have *semantic content*, and syntax is insufficient for the emergence of meaning. In arguing thus, he understands the distinction between syntax and semantics as the conceptual truth of our distinction between the notion of the purely formal and that which has content (see Searle 1984). Intuitively, this may be acceptable and even expedient for distinguishing different aspects of language but if we are looking for a way of possibly moving from syntax to semantics, Searle's assumption of a complete gap between syntax and semantics is unjustified and artificial. Some authors have therefore recently pointed out that disjunction in such a case does not make good sense (e.g. Peregrin 2021), or have tried to show from the beginning that syntax is sufficient for semantics (Rapaport 1995; 2002).

---

[3] It must be said that Searle, in his responses to Dennett's *Consciousness Explained*, was confronted with the advantages of parallel computer architecture, virtualization, neural networks and their ability to learn, in general, the position of connectionism, without this leading him to reconsider at least some of his original arguments.

Yet Searle is not alone in arguing for a deep divide between syntax and semantics. Colin McGinn explains this gap in a similar way when he argues that symbols can be manipulated on the basis of their meaning or just on the basis of their syntactic form. The syntactic manipulation of symbols is guided in the execution of operational rules only by the symbols themselves and not by what they mean. A computer program, then, is a symbol manipulation algorithm that mechanically applies manipulation rules to achieve a desired result. Thus, the computer does not have to deal with the meaning, but only follows the syntax (McGinn 1999, 180). Although McGinn does not draw conclusions as radical as Searle's from the dissociation of syntax and semantics—he does not claim, for example, that *syntax is insufficient for semantics*—he similarly insists on the fundamental difference between brain and computer processes and does not assume that meanings play any role in computer information processing. "My mental processes involve the manipulation of meanings, not merely strings of syntax. I am a semantic manipulator, as well as a syntactic one." (McGinn 1999, 182) A computer, unlike mental processes, is merely a syntactic manipulator "because it has been cunningly programmed to manipulate symbols in a way that mimics understanding." (McGinn 1999, 181) Thus, McGinn does not assume that a machine can understand language, but merely simulates understanding, and with this disjunction between simulation and replication McGinn is thus subscribing to Searle's position.

Although many computer programs, especially within the symbolic approach, can be characterized in the way McGinn describes, I argue that even in this case we accept *mechanical symbolic manipulation* without sufficiently considered assumptions.

First: Even the mere mechanical manipulation of symbols requires the identification of the symbol as a symbol and its position in a chain of other symbols, which has implications for the application of this or that rule. Let us consider these as *minimal semantic contents* that are a necessary prerequisite for a computer to follow the syntax, so to speak. Thus we cannot say that the computer need not concern itself with meaning, but we could say that the computer need not concern itself with their more complex possible meanings. However, some minimal semantic contents is a condition for syntactic manipulation to be feasible. I discuss a concrete example below.

Second: If a computer program is written to manipulate symbols primarily on the basis of their minimal semantic contents, this does not preclude the possibility that the computer program could be constructed in a different way that would not only respect these minimal semantic symbolic manipulations, but that its design would take advantage of a different ordering and processing of information within a neural approach and use more complex meanings. How this can be implemented in current LLMs will be presented next.

The minimal semantic content of syntactic entities and the possibility to follow more complex semantic content in symbolic manipulations causes the thesis of the abysmal separation of syntax and semantics to be judged by intuitions that are not sufficiently justified. In a similar vein, Dale Jaquette has challenged this seemingly abysmal separation of syntax and semantics (1990): "If we mean by pure syntax, ..., syntax entirely divorced from semantics, then I think there simply is no such thing. Without at least some semantic content, what is sometimes loosely referred to as pure syntax is nothing but marks on paper or magnetic patterns on plastic disks." (Jacquette 1990, 294)

The interconnectedness of syntax and semantics was not denied later by Searle himself, who wrote, "And indeed, though it is less obvious, I think it is also true that an entity can only have a syntactical interpretation if it also has a semantic interpretation, because the symbols and marks are syntactical elements only relative to some meaning they have. Symbols have to symbolise something and sentences have to mean something. Symbols and sentences are indeed syntactical entities, but the syntactical interpretation requires a semantics." (Searle 2002, 117) In this quotation, Searle is undoubtedly expressing the interdependence of syntax and semantics that we are trying to justify here. The fact that this did not become a reason for Searle to reconsider his original thesis and possibly further revise his conclusions can be explained by the new form of his *Chinese room argument* (Searle 1990) where he attributes to syntax and semantics only epistemological objectivity and full dependence on the observer.[4]

## 2.1  Syntax implies semantics

The premise *syntax is not sufficient for semantics* is not so justified that we cannot abandon it. Let us now propose another possible premise for the relation between syntax and semantics: *Syntax as a formal principle presupposes at least a minimal semantic content of the entities it concerns.*

Consider the elementary character system of Morse code. Let us leave aside for the moment that this is a way of encoding the letters of the alphabet; it shall suffice to distinguish the basic entities of this system, i.e. dots and dashes and their possible ordered combinations. *Syntax* as a formal principle is possible in this case only on the assumption that the (*minimal semantic*) contents of the individual elements, i.e., the dot and the dash, are distinguishable, and that the ordered combinations of such elements are distinguishable as either identical (e.g., as three dots in a row and three dots in a row) or different (e.g., as three dots in a row and three dashes in a row), and similarly other possible combinations of dashes and dots.[5] If we did not allow for the existence of these minimal semantic content of entities, they would not be distinguishable from one another, and neither would be their ordered combinations. In such a case, there could not even be any formal syntax for these entities. This again, as in the example above, shows quite convincingly that syntax cannot exist independently of semantics: it always requires at least some minimal semantic content of entities.

However, in this case we could face the objection that the semantic content should not be associated with the meaning of the sign as a sign, but if it is to be semantics, then the sign must symbolize or refer to something different, something outside the symbol itself. The thesis does not deny this requirement, but instead looks for assumptions about how a symbol can refer to something different. Consider, therefore, the following rule: *A symbol must first refer to itself in order to refer to something distinct from itself*. Self-reference is a condition for its distinctiveness in some syntactic arrangement, and its other more complex possible references are determined by the wider context. If, for example, it is the symbol of the three dots in Morse code, then self-reference guarantees the distinguishability of the sign in a stream of many distinct or similar signs, and it can also symbolize the letter *S,* as a more complex meaning given by the assignment of the signs and letters of the alphabet. In this elementary case, the reference to a more complex meaning is simple because it is determined by the uniqueness of the symbol and letter assignment. In other cases, however, the complexity of possible references may be infinite and is then limited only by the surrounding context. For example, the sign or symbol "PRAGUE" refers in general to an infinite number of possible different descriptions, which are not and cannot be currently considered, but are in each particular case reduced, better or worse, by the context of the given meaning under consideration.

---

[4] The problem of the objectivity of syntax and semantics is dealt with elsewhere.
[5] In this example, I do not consider the separator as a necessary part of the entities, because it can be replaced in any way that distinguishes the ordered combinations clearly enough.

The above thesis regarding the minimal semantic content of the entities to which syntax refers thus challenges Searle's thesis that *syntax is insufficient for semantics*, and argues instead that *syntax does not exist without semantics, and that the fact of their mutual conditioning makes it possible to move from minimal semantic contents to semantics of more complex contents*. Or in other words, that *respect for syntax is a precondition for the emergence of other higher semantic contents*, or alternatively, the other way around, that the *distinguishability of differently complex semantic contents is a precondition for the distinguishability of their "orderliness", i.e. syntax*.

The fact that syntax does not exist without semantics has played a crucial role in other approaches to the interpretation of formal symbolic systems. The fundamental question in these approaches, however, is not so much the link between syntax and semantics, but rather the answer to the question, how are meanings grounded in symbols? I shall briefly mention some crucial ideas of Harnad's "the symbol grounding problem" (1990) and Rapaport's "syntactic semantics" (1995) as prerequisites of a more convincing interpretation for understanding LLMs.

## 3 Symbol grounding problem

Stevan Harnad (1990) attempted a semantic interpretation of the formal system of symbols without having to refer only to the meanings in our heads. Similar to my above emphasis on the minimal semantic content of entities that can be distinguished and identified, Harnad attends to "discrimination and identification" to emphasize the elementary ability in our perception to *distinguish* between the same and different inputs, and for different inputs to *identify themselves*, i.e., how much they differ (Harnad 1990, 341). The ability to discriminate and identify is linked to *iconic* and *categorical* representations, in which sensory analogue transformations of the projections of distal objects on our sensory surfaces produce their iconic representations and selectively reduce them to categorical representations. Both iconic and categorical representations are still sensory and non-symbolic (Harnad 1990, 342), but they ground a certain set of elementary symbols. Once this basic set of elementary symbols is available, "the rest of the symbol strings of a natural language can be generated by symbol composition alone." (Harnad 1990, 343) Harnad thus sought to propose a collaboration between *symbolic* and *connectionist* approaches to modelling the mind. A purely autonomous symbolic system is ungrounded: its symbols do not refer to any objects. The purely connectionist model does link names to objects, but it lacks compositional ability. Thus, the network of names is already grounded but is not amenable to a full systematic semantic interpretation. The task of connectionism has been to provide a mechanism of categorical representation so that the manipulation of symbols is governed not only by the arbitrary shapes of the symbols but also by the non-arbitrary shapes of icons and categorical invariants, i.e. their meaning. Thus, meanings are not grounded in our intentionality but may very well be grounded in other symbols. This conclusion is crucial from the point of view of language models, and I shall return to it at the end of the paper.

### 3.1 Syntactic semantics

A fundamental disagreement with Searle's supposed intuition about the insufficiency of syntax for semantics was expressed by William J. Rapaport (1995; 2002). Syntax alone is sufficient for semantics, and although it may seem somewhat paradoxical, "semantic understanding is syntactic understanding" (Rapaport 1995, 49). The argument runs briefly as follows: if semantic understanding is based on a relation between two domains, whereby I understand one domain through the terms of the other, then each domain can be understood recursively through the terms of the other domain. However, such a procedure cannot be infinite, and there must be a domain that is not understood in terms of the other domain, but simply through the terms of the domain itself (Rapaport 1995, 49). In the case we have presented, we speak of the necessary *self-reference* of the entities of a symbolic system as a necessary assumption for the existence of more complex references which depend on the context. Rapaport similarly assumes that the initial domain or area is not grounded in another domain, but grounds itself as its starting point. Rapaport also refers to Harnad and his basic set of symbols in this regard, finding that although we would expect internal entities to be grounded in external entities, we find that it is the *internal representations* of external objects in the mind that are the referents for the grounding of other internal symbols. Since these are all internal representations, the meaning is in the head and is syntactic (Rapaport 1995, 80). Again, then, not only is syntax closely tied to semantics, but semantics itself becomes syntactic semantics. All understanding rests on syntactic understanding.

## 3.2 Vector grounding problem

Recently Mollo and Millière (2023) have built on Harnad's and Rapaport's efforts to assign meanings to entities of a symbolic system and have attempted to find such a solution in the case of ANNs and LLMs, which work with vectors rather than individual symbols. The difference may seem significant because in the case of symbolic systems we are used to thinking of symbols as basic units of meaning and strings of symbols as more complex meanings, whereas a vector-based, numeric representation may seem a much more complex entity not only in terms of meaning but also in terms of its ability to hold subtle distinctions in its occurrence within a given context of larger textual entities. In reality, however, this difference is not significant when we consider the grounding of these representations. The essence of the grounding problem does not change if we work with continuous vectors rather than discrete symbols "when it comes to escaping the merry-go-round of representations and being connected to the external world" (Mollo and Millière 2023, 7).

Furthermore, it is in fact very difficult to establish any representational level at which we want to ascribe or deny meanings to some elementary symbols or vectors. In the example of Morse code, we certainly tend to think of the elementary symbols of dots and dashes and their combinations. In reality, however, dots and dashes are represented in some way, e.g. in the 8-bit system by the number of 0s and 1s in a single byte, e.g. 00101100. Yet even this level is symbolic, because it does not actually contain any 1s and 0s, just as 00101100 is not a dot. The 1s and 0s are represented by different voltage ranges (e.g., 0 in the range 0-0.8V and 1 in the range 5-5.8V) on the microprocessor pins. We could also descend to deeper representations and look for the cause of different voltage states in the structure and the number of electrons and protons. Perhaps even such a representation is not the final possible level, nor is there any reason to claim that the above are all of the possible symbolic representations. So the question is, how can an internal representation have not only a minimal semantic meaning (necessary for its distinguishability and identification) but also a much more complex meaning tied to something outside the representation itself? Mollo and Millière argue that of the five conceptually distinguishable ways of grounding internal representations in biological or artificial systems, *referential grounding* is the central and most important because, through historical-causal relations, it allows some LLMs trained only on linguistic data to achieve the desired relevant form of grounding (Mollo and Millière 2023, 2).

In doing so, the authors assume that they can solve the so-called "vector grounding problem" without committing to answering questions about whether LLMs can "understand language (or anything at all), perform linguistic acts, or have cognitive or mental properties" (Mollo and Millière 2023, 2). Thus, they assume that these questions are irrelevant and that the way in which they can achieve grounding vectors in the real world, with which LLMs' language models have no direct causal contact, is fully sufficient to explain the achievements of generative AI. In doing so, they shield the deepening of referential grounding by tuning language models in processes called Reinforcement Learning from Human Feedback (RLHF), although they admit that there are language models that can achieve referential grounding in limited domains without RLHF (Mollo and Millière 2023, 28).

However, what does the successful referential grounding of LLMs get us if it is not to serve as evidence of language *understanding*? If the referential grounding of LLMs is only meant to be a justification for their success, then the question creeps in: could LLMs not be just as successful without referential grounding? Finally, if language understanding is irrelevant if the language model is grounded in the world, then one has no choice but to assume that language understanding must be tied to cognitive-mental states, to awareness of understanding. Awareness, grounding and understanding are, in fact, the vertices of an imaginary triangle whose different relational conditionings determine the interpretative position that different authors take.

Searle (1984) requires as a condition of understanding not only a grounding of language in the world but also an awareness of understanding - an intentional state that is about something. Dennett (1987) and Søgaard (2022), on the other hand, assume that an awareness of understanding is not a condition for understanding language because, even if understanding happens consciously, empirical cases of unconscious understanding can be documented. Mollo and Millière are then reluctant to say whether or not LLMs understand language, but they show that at least some of them achieve referential grounding in the world. Harnad seeks a way to rid a purely symbolic system of its lack of grounding, without the symbols having to be parasitic on the meanings of the mental states of the performer, but instead inherent in the symbol system itself. In this case, referential grounding is already significantly weakened, but it is still somehow mediated by analogical projections of external objects to secure the meanings of the underlying symbol set. Rapaport then notes that although we would expect all internal symbols to be "grounded" in external objects, the grounding seems to be limited to symbolic representations and elementary symbols, which are all *internal*. So the question arises, is the prevailing tendency towards the *referential* grounding of symbolic and vector AI systems justified?

## 3.3 The myth of referential grounding

The referential grounding of the meaning of symbols is often considered a central and decisive way of assigning meanings to purely formal symbols, even when it comes to grounding vector-represented tokens as the elementary parts of meanings at a chosen level of resolution. The underlying assumption of referential grounding is that the bearer of such a symbolic system is causally-historically embedded in the external world and the symbolic system represents this external world. According to many, the *pointing game* is a litmus test of the grounding of such a symbolic system (Søgaard 2022, 442). An ungrounded system, or a system showing only partial grounding in the world, would not be able to respond well enough and would 'flounder', for example in those cases that require grounding through ostensive definitions of the meanings of some (or even all) symbols. It is easy to imagine that the bearer of such an ungrounded or a poorly grounded system will not be a successful agent, i.e., if they need to adapt their behaviour to causal-historical influences in the world.

However, such an idea is too simplistic if such a symbolic system is to be understood as a language relating to the external world, as a language that represents the world. Of the many problems associated with how such a system might represent the world, let us consider the following. A simple symbolic system requires the establishment of some rules for operations on symbols, which are usually defined outside the symbolic system in some more complex system, a metalanguage. Thus, the fact that a symbol can represent something is then a matter not only of that symbol system, but also of the rules established. Yet to what extent can something be represented by the symbol itself, and to what extent is this representation then dependent not only on the symbol and the rules set, but also on the metalanguage in which the rules are formulated? Whether or not one can determine the extent of this dependence, in the case of the LLMs of generative AI such a difficulty falls away. Although some authors (e.g. Mollo and Millière 2023) call for grounding these language models as well, such a requirement can be considered a redundant and unfeasible myth.

Referential grounding is problematic in itself, even in the case of a symbolic approach to AI, where we would only require the referential grounding of individual symbols. Trying to ground the meaning of a symbol (e.g., the word *cat*) in the "outside world" will be as successful as trying to add one real cat to every dictionary that contains a definition of the meaning of such a symbol through strings of other symbols (Rapaport 1995, 78). Ultimately, then, to each dictionary one must also add a version of the real world in which the system is to be grounded. This is reminiscent of modern empiricism's attempt to follow Fries' solution to the trilemma between dogmatism, infinite regress and psychologism in epistemology. As Popper has already shown ([1935] 2002), in Fries' example, if we reject dogmatism as unjustified knowledge, we are left with the infinite regress of logically justifying a given proposition by means of other propositions, or the "stepping aside" to our sensory experience, in which, as Fries supposed, "we have 'immediate knowledge': by this immediate knowledge, we may justify our 'mediate knowledge' - knowledge expressed in the symbolism of some language." (Popper 2002, 75) Referential grounding assumes that one can escape circular (and in this sense infinite) dictionary definitions, where individual symbols are defined through strings of other symbols, by taking a "step aside" and looking at the actual *cat* supplied with the dictionary. Our actual sensory experience of the real *cat*, and gradually then of the rest of the real world, should ground the meanings of the language system - but such a grounding is illusory. What are the reasons for this illusion?

On closer analysis of Harnad's (1990) essay, Rapaport notes that "the relation between word and world is really a connection between an internal representation of a perceived word and an internal representation of the perceived world", and that symbol grounding "does not necessarily get us out of the circle of words - at best, it widens the circle." (Rapaport 1995, 78-79) Language is thus not directly connected to the external world, but rather to its linguistic articulation (the internal conceptual representation of the world), because everything that is imaginable must also be linguistically articulable; the boundaries of language thus define the boundaries of the world (Wittgenstein 1974 [1921]). Why then, in such a case, consider the "pointing game" as a test of machine understanding of language? Does the act of "pointing" ("this is a cat") assign meaning to the symbol *cat*?

Harnad assumes that iconic and categorical representations, i.e. sensory and non-symbolic representations that allow discrimination and identification, take place at the moment of "pointing". The categorical representation is then the result of the *selection* of the iconic representation so as to "reliably distinguish a member of a category from any nonmembers with which it could be confused" (Harnad 1990, 342). Although such a mechanism is rather obscure in its partial implications, Harnad admits that even if some representations are innate, in other cases these representations may be established in the course of experience, and thus categorical representations must necessarily undergo gradual change. Thus, if we ignore the fact that referential grounding does not ground words in the world, but to an internal representation of the world, then in the act of "pointing", e.g., "this is a cat", there is not only a reliable selection of a category member from other members, but also an establishment of the category relative to the one being pointed to. Consequently, such grounding is illusory because what the meaning is grounded in is continually changing. Otherwise, we would have to assume that there is either an initial phase during which the categorical representation is merely being formed or is established in a one-off way by some type of baptismal act. Both of these possibilities are unlikely, however, because no specific cases can be cited in their favour, nor do we have any internal experience of such ways of dealing with categorical representation. They are used as if they were given from the beginning in a way that expects them to be continuously transformed and shaped. Grounding in the ungroundedly changing relative to the grounded is thus illusory.

Even more fatal to the possibility of referential grounding are Popper's objections to possible justification on the basis of "immediate experience", for every proposition is inherently transcendent because it contains universal terms (names, symbols, ideas) that transcend our immediate experience. Therefore, the "universals which appear in it cannot be correlated with any specific sense-experience. (An 'immediate experience' is only once 'immediately given'; it is unique.) ... Universals cannot be reduced to classes of experiences; they cannot be 'constituted'." (Popper 2002, 76) This ultimately means that universal names or symbols cannot be constituted by immediate experience *at all*. The indicative statement "this is a cat" not only contains the universal term "cat," which transcends the act of an immediately given experience of perceiving an object, but also fails to import meaning to such a term in any appropriate way. What relation "to the external world" or "of the external world" should be established by this pointing act? A pointing statement of the type "this is x" cannot constitute universals, cannot continuously shape the meaning of symbols or words, nor can it assign meaning at its first use, in a kind of analogous "baptizing" act. The referential grounding of meanings is thus an illusion and does not actually provide the symbols with the desired grounding of meaning.

I have given more detailed attention to referential grounding because it is a frequent topic of consideration in papers discussing the performance of generative AI that turn to reference theory as a traditional semantic theory. However, referential grounding is not the only way to fix meanings for a symbolic (linguistic) system. Yet the diverse ways of conceiving meaning in semantic theories are so different, and so lacking overwhelming consensus, that it is difficult to give any one way priority on the basis of theoretical argument. The fact is, however, that most of these conceptions have been developed with the implicit conviction that the bearer of natural language is only a conscious agent, a human being aware of their relation to the world in the predicate of their intentional stances. Now, however, all semantic conceptions must somehow come to terms with the situation in which a machine is an equally successful bearer of natural language.

### 3.4 Alternative concepts of semantics

In addition to the prevailing tradition of *referential semantics*, a rather broad alternative stream[6] of *inferential semantics* has emerged which, following Wittgenstein, assumes that what matters in terms of linguistic meanings is rather the "use of the linguistic expression" within the *inferential roles* of the rules that govern the use of the expression. Inferential role semantics (IRS) is part of a more general stream of *conceptual semantics*, which "takes the meanings of words and sentences to be structures in the minds of language users, and it takes phrases to refer not to the world per se, but rather to the world as conceptualized by language users." (Jackendoff 2019, 86) This shift from the external world to an internally represented conception of the world has already been highlighted above in relation to the attempt to address the relationship between syntax and semantics in both Harnad and Rapaport. Because conceptual semantics weakens the direct linking of meanings to the external world, it seems a more appropriate theoretical basis for interpreting the grounding of meanings in language models. Harnad's proposal, however, assumed a combined use of symbolic and neural approaches, with neural networks counted on only to provide iconic and categorical representation, and the language model built in a standard symbolic way. However, current LLMs (transformers) are substantially different and their functionality in language processing is necessarily dependent upon the structure of neural networks. Also, the requirement not to refer, in terms of meanings, to the world as such, but to the conceptualized world in the minds of language users, is only a weakening of direct reference to the world but not a sufficient explanation for grounding the meanings of language models. The grounding of meanings within conceptual semantics works with some form of internal representation of the world that is conscious to the language bearer. As we have already seen, however, consciousness playing such a role is not necessary in order to understand language, nor can we yet attribute it to a machine. Grounding the meanings of language models must therefore be done in a different way.

The fact that even inferentialism requires grounding meanings in some interaction with the world is understandable in terms of the representational role of language. Global inferentialists, for example, assume that it is necessary to construct inferential rules very broadly to include perceptions as "inputs" to language and intentional actions as "outputs" from language. Then the pointing statement "this is a cat" should secure the link between perceptual experience and the representational role of language, and likewise for the link between linguistic expressions and subsequently motivated action. Inferentialists worry that "[I]n the absence of any such grounding of meaning in our experience and interaction with the world, our inferential language game threatens to fail to 'latch onto the world', it could not serve as a means for representing the world" (Murzi and Steinberger 2017, 6).

---

[6] Terminologically, the individual inferential paradigms of semantics are referred to differently. For example, the *conceptualist* or *cognitivist* tradition of semantics is thus defined in relation to the traditional *realist* or *externalist* (Gärdenfors 1993); inferentialism or inferential role semantics (Brandom 1994; Peregrin 2014) is a special case of a more general stream of conceptual semantics (Jackendoff 2019), functional semantics, procedural semantics (W. A. Woods 1981), and dynamic semantics (e.g. Kamp 1981, Heim 1983).

If we have already rejected the possibility that the pointing game in the world has any direct influence on the constitution or grounding of meanings, why should we count it within the broader rules of inferentialism? It makes no difference, after all, whether the pointing "this is a cat" is intended to be *primarily* truth-bearing in terms of referential semantics or to follow *primarily* its inferential role, the rule that establishes meaning. The demonstration "this is a cat" can be a true or false statement depending on what I am pointing at, but it can also be a normative rule, the use of which will put me in agreement with other language carriers if our experience agrees that it is a cat, or in disagreement if it is not a cat. However, neither of these options directly establishes the meaning of the *cat* symbol. *The pointing game thus has a primarily correlative and synchronizing role. It allows us to correlate the linguistic system with the world and to synchronize individual linguistic expressions into larger formations so that they fulfil their linguistic roles*.

As far as the current LLMs of generative AI are concerned, it is clear that we cannot attribute to them an effort to "attach to the world" linguistically. Hence, the prevailing view is that language models only operate stochastically with linguistic expressions, without regard to their actual grounding in the world which, as we have seen, ultimately means that they cannot understand language but only simulate their understanding. Inferentialism, however, is a decidedly more plausible explanation of the workings of language, as opposed to the referential grounding discussed above, because it assumes that the meaning of a linguistic expression is determined by the rules of its *use* in the language game. Since language models lack referential grounding but appropriately use linguistic expressions in language games, it can be inferred that referential grounding is not essential to language functioning but may be *advantageous* for some language speakers, for whom it allows better orientation and interaction with the world. Not surprisingly, then, such a practice leads to an overemphasized referential conception of meaning, or to a belief that it is important to maintain at least broadly conceived inputs and outputs "from language" in order to preserve the link between perceptions and representations of language and thus maintain the "attachment" of language to the world.

What is crucial in this language users' game is primarily the successful linguistic inputs and outputs, which of course removes the possibility of a more detailed recognition of the linguistic patterns hidden between inputs and outputs. As the capacity for linguistic abstractions and generalizations gradually increases, these hidden linguistic mechanisms become more and more inaccessible to organic language bearers. Yet how is it possible that non-organic language bearers, who do not directly connect language to the external world referentially or inferentially, i.e. the LLMs of generative AI, are still able to successfully participate in language games and work with meaning?

## 3.5 Unit of meaning

The problems of grounding the meanings of symbols at the level of words, discussed above, assumed the adoption of semantic atomism, which considers words as elementary units of meaning, the meanings of larger linguistic expressions (e.g. sentences) then being determined by the composition of meanings of linearly ordered words. The assumption of such a *principle of compositionality* is that the meaning of a complex expression is determined by its structure and the meaning of its individual components. Although such a principle may seem intuitively unproblematic, its implications gradually lead us to the conclusion that it should be possible to compare the truth of the individual components of more complex expressions with our experience and thus decide on their truth or falsity. This dogma of reductionism has been challenged, with reference to Carnap, by Quine's famous statement that "our statements about the external world face the tribunal of sense experience not individually but only as a corporate body." (Quine 1951, 38) What implications and possibilities for exploration are offered by this reversal of reductionist assumptions and by the holistic approach to grounding the meanings of linguistic expressions in models of generative AI?

## 3.6 Semantic atomism and molecularism

Without assuming that we can find a safe path in the somewhat wild landscape of atomistic, molecular, and fully holistic conceptions of meaning, I shall at least try to suggest a direction in which the desired explanations might be approached. Most contemporary semantic theories of meaning are holistic, though not always in the radical way that Quine (1951) and Hempel (1950) proposed. Brandom's inferentialism, for example, subscribes to *semantic molecularism* (see Dummett 1991) where the focus is on *sentences* as the basic units of semantic explanation, thereby defining itself in relation to both *semantic atomism* and fully *holistic* theories (see e.g. Brandom 2007, 671). "As it is only sentences that may be used to make a move in a language game, any contact between a word and (a part of) the world must be mediated by sentences." (Peregrin 2014, 31) Our familiar sentence "this is a cat" must therefore be read in terms of inferentialism as a basic unit of meaning, a broadly defined inference that is an "entry into" language from experience in the situation of showing a cat in the world. Thus, the basic semantic unit is the sentence as a whole in relation to the world, not the individual isolated word. However, semantic molecularism does not assume that it is possible to understand one isolated sentence without understanding others. Dummett argues that the principle of compositionality allows us to distinguish between *representative sentences*, in whose context a given word acquires meaning, and then derived sentences, in which we understand the word but which are not required to understand the word (Dummett 1991, 224). The meaning of words is thus contextually bound to a certain range of sentences in the language in which the word occurs. This seems remarkable given the way language models work with the meaning of linguistic expressions (tokens) and their mapping to multidimensional vectors. The unsurprising fact is that Frege's *Context Principle* applies here (see e.g. Stainton 2005) as a condition for the successful determination of the meaning of a linguistic expression. At the same time, the gradual extension of context is also a crucial step for the success of individual language model architectures in NLP.

The paradigmatic advantage of transformers is precisely the way they are able to work with context. Previous methods have attempted to capture the varying context of a word (more generally, a token) by storing the relations of a given word to surrounding words in varying degrees. This created a language model in which each word was defined by a multidimensional vector of values into which the relationships of words to surrounding words were mapped. The generated dictionary stored the semantic and syntactic relations between words, and the vectors of each word could then be compared to determine their semantic and grammatical proximity, similarity or contrast. The resulting dictionary was therefore able to capture *word-embeddings*, but only by static methods (e.g. Word2Vec or Glove), which did not allow for the distinguishing of the different meanings of words possible in given contexts. The first innovation for capturing *contextualized word-embeddings* ELMo (Embeddings from Language Models) (Peters et al. 2018) had already been implemented in the framework of RNN/LSTM (Recurrent neural network/Long-short term memory models) methods, which are now outperformed by transformer models originally developed for language translation. However, both parts of the transformer architecture, i.e. encoder and decoder, are now successfully used to train language models with contextualized embedding. GPT (Generative Pre-trained Transformer) uses the decoder part and BERT (Bidirectional Encoder Representations from Transformers) uses the encoder part. In both cases, the resultant trained language models can then be used for various NLP tasks.

Could we then assume that some form of inferentialism is sufficient to explain the success of language models in processing natural language? We have seen that Brandom's inferentialism invokes Dummett's semantic molecularism, in which sentences play a crucial role as basic semantic units. However, if context is crucial for determining the meanings of simpler linguistic expressions, then not only the word but also the sentence may still not be a sufficient extension for the basic unit of meaning. Indeed, there are examples of identical sentences that have different meanings in different contexts, and it can be assumed that there are larger textual formations whose meaning is similarly contextual. It seems, therefore, that it would be more appropriate to think of language holistically, as a whole that also addresses the contextuality of larger textual formations. Although Dummett rejects any form of linguistic holism and prefers a compositional theory of meaning, he acknowledges that the meaning of a word is not given atomically but is only discernible from a certain range of the representative sentences mentioned above. The principle of holism thus enters semantic molecularism by the back door, if we consider that if the condition for understanding the meaning of a word is some range of representative sentences in which that word occurs, then this necessarily implies understanding the meanings of the other words in those sentences. However, their meaning is bound by the different range of representative sentences for those words, and similarly for the other words in that range of representative sentences, and so on, until the language as a whole is connected in this way. Dummett makes the understanding of a sentence conditional on a fragment of language that sufficiently grounds the meaning, but since the scope of such a fragment cannot simply be determined, it cannot be ruled out that this fragment must ultimately be the language as a whole. Dummett claims, "To understand a sentence of a given language, one must know some fragment of that language, in which, of course, much would be incapable of being expressed, but which could in principle constitute an entire language." (Dummett 1991, 222) Can we somehow support these ideas by processing meaning within the LLMs? How large would such a fragmented contextual unit of meaning be, and how might meanings be grounded in such a framework?

## 3.7 Semantic fragmentism

In the context of the functioning of language models of generative AI, it seems that we could consider a minimal unit of meaning to be a *linguistic corpus*, i.e. a *fragment* of language, a linguistic unit that is contextually interconnected. Terminologically, corpus seems to be a more appropriate term than, for example, *discourse*, which is also understood as a larger linguistic unit connected by meaning. Corpus is, however, a much more technical term in this respect, and really refers only to a linguistic body, a semantic fragment, a slice or somehow semantically bounded area of language. The analysis of language corpora is a common practice in computational linguistics and allows for the empirical testing of established hypotheses of various types. From a philosophical point of view, however, a corpus should be understood as a certain subset of language as a whole and, therefore, a part of the world that is defined by the boundaries of a given language corpus. Theoretically, then, we can understand the whole of a language as a corpus of all language corpora and the boundaries of the world defined by the boundaries of this all-encompassing corpus.

Why do we choose such a somewhat clumsy path to the whole of language? The reason lies in the difficulties we would face if we were to commit ourselves directly to a linguistic holism that would identify the unit of meaning at the level of the language as a whole. A holistic conception is in this case tricky when it comes to defining the whole and its boundaries. From a holistic point of view, the whole cannot be determined by the sum of all its parts and, in the case of language, therefore by the sum of the meanings of all the more elementary linguistic entities, i.e. words (atomism) and sentences (molecularism), because not only does their mere *sum* not provide a whole in the holistic sense, but in the case of language the *sum* of all words and all sentences does not make good sense either. The totals of possible words and sentences are infinite not only because new words appear, but also because an infinite number of sentences can be generated from a finite number of words.[7] Difficulties of this type, however, are not now deal-breakers and their resolution is not a condition for understanding how linguistic meanings are grounded within LLMs.

---

[7] In language there is no limit to the number of repetitions of a particular word in one sentence.

The basic unit of meaning is thus related to the fragment or corpus of the language as a whole (whether it is an isolated corpus or a corpus of all corpora) and not to individual sentences or words. That words, whole sentences, and perhaps even larger sections of texts can have different meanings in different contexts perhaps needs no proof. It is therefore necessary to consider a particular linguistic fragment or corpus as a unit of meaning. Assume that it has a minimum size in the sense that it sufficiently determines the meaning of its individual parts, i.e. sentences and words.[8] *Minimal size* means that such a corpus is minimal in the sense that additional variations of linguistic expressions can be added to the corpus without changing the grounding of the meanings of the individual parts of the corpus, i.e. sentences and words. Thus, a minimal corpus is semantically saturated and is sufficient to ground the sentences and words it contains. If, however, there are linguistic expressions that necessarily belong to the corpus in terms of the context and their inclusion in the corpus changes the grounding of the parts of the corpus, then the initial corpus is not minimal and is not semantically saturated. These complex semantic relations of the individual parts of the corpus to all other parts of the corpus then ground the meaning of the isolated parts in a given context. The outlined structure of grounding meanings within language models should not be understood methodologically as a practical procedure for determining meaning saturation and thus minimum corpus size, but rather as a theoretical description of how meanings are grounded within language models.

## 3.8 Grounding the meanings of LLMs

The meanings of linguistic expressions are grounded neither in the world, nor in an internal idea of the world, but in the linguistic corpus as a whole. The holistic role of language implies that the meanings of individual linguistic expressions are grounded in multiple relations with respect to other linguistic expressions and their combinations. Such a linguistic model does not require the referential grounding of individual linguistic expressions in extra-linguistic existing entities or at least in the internal representations of such entities. Language as a whole is constituted by basic, identifiable entities whose meaning is determined by the structure and frequency of relations to other entities in the context of linear sequences of such entities, i.e. sentences. Within such a model, the requirement that meaning be determined by the "use of an expression" in the language game remains intelligible, since its use takes into account not only its relations to previous and, predictively, to possible future expressions, but also to all other expressions in relation to which the expression in question grounds its meaning. Thus, the meaning of an expression is not something essential that lies behind the expression or to which the expression refers. Such a delineation of meaning may seem implausible and paradoxical, since the meanings of an individual expression do not exist without the complexity of its relations to other expressions, but whose meanings are again determined only by similar other complexities of relations to other expressions. *If an expression is to represent something, then it represents its place in the complexity of relations to all other expressions.*

In response to this model, the following three objections are offered. Firstly, if we take an isolated linguistic expression, e.g. *cat*, we immediately associate with it its most frequent and likely meaning, without being aware of any complexity of relations to other expressions. So how do we explain that an isolated expression has meaning? The answer is that it is not possible to isolate individual linguistic expressions from language. They can be put into a kind of meaning indeterminacy in terms of context, but they cannot be fully isolated from the system of relations of language as a whole. Thus, we attribute a meaning to the expression *cat* depending on the context, which we do not know due to isolation, but we assume the most likely meaning, or several possible meanings, given our linguistic experience. The unacknowledged complexity of relations to other terms is a consequence of our focusing only on linguistic inputs and outputs. The complexity of the patterns of such expression relations is too complex to be practically and usefully reflected upon and to provide guidance in language use. Meanwhile, this complex pattern of language relations is processed in hidden layers of the neural structures of our brain, similar to the hidden neural layers of language models, while only the input and output layers are consciously accessible to us. However, this does not mean that these complex language patterns do not exist and are not crucial at appropriate levels for the functioning of language.

---

[8] Strictly speaking, we should list tokens, but this does not in principle change the grounding of the expressions of the individual parts of the corpus.

Another possible objection is that if none of the linguistic expressions have an essential content, where can any meanings appear at all? The answer is that the meaning of a linguistic expression is nothing other than the set of relations to other expressions. By such a set we mean the sum of all syntactic, semantic and other linguistic "patterns" that exist in the language and that we unconsciously process between the inputs and outputs of the language, and that are similarly recognized and subsequently stored within the NLP of LLMs. If we play the "pointing game" when the linguistic expression "this is a cat" is used, then we may come to believe that some particular cat we are pointing at, or our sensory perception of a cat, grounds the meaning of the linguistic expression "cat" to which the linguistic expression refers; but in fact we are only reifying a place for the linguistic expression "cat" in the overall complexity of the relational expressions of language. The very act of pointing then appears to be merely a convenient correlative matter between language and the empirical world.

A third possible objection is that such a linguistic model deriving the meanings of linguistic expressions only from the complex relations of the expression to other expressions would be too fragile to maintain its consistency and would fall apart at the first possible modification. For example, a relation could be added or removed due to the creation of a new linear sequence of linguistic expressions (sentences) or the recognition of a misplaced expression in the vicinity of other expressions. The answer is that shared language is not static and dynamically changes just like that all the time, while language models are used thus only with tuned fixed values of weights and biases for each hidden layer of the neural network model. Neither natural language nor fixed language models break down as a result of changes, but exhibit a sufficient degree of plasticity to accommodate the necessary modifications to complex sessions. The fact is that language models are sensitive to the data on which they learn, and errors in the data significantly affect their functionality.

If language does function as a complexity in which meanings are given by the interrelations of more elementary expressions, it is not impossible that sufficiently complex artificial structures can play with language as successfully as conscious agents, such as humans. The fact that one can express in statistical terms the frequency of a linguistic expression in a given context and in relation to other linguistic expressions does not mean that one is guided by these statistics in communicating language, nor does it mean that language models using stochastics to fill in previously masked words are mere repeaters guided only by statistics and probability. The addition of another word in a linearly constructed sequence of linguistic expressions, taking into account its probability with respect to the other parameters of the model, is not all that determines the generated form of the text. Above all, it is a sufficiently large context that significantly determines the generated text and the consideration of those linguistic patterns recognized in the language that nonlinearly influence the content when processed in the hidden layers of the network. It would be naive to think that only the probabilistic repetition of patterns can lead to the linguistic creativity and originality achieved.

As consciousness-endowed language bearers, we believe that we are guided in language processing by the pursuit of certain intentions, which we express in linguistic expressions. Yet how can a mechanism that translates a given intention into a linear sequence of linguistic expressions work? Whatever the nature of the intention, if it is to be perceived and understood as an *intention*, then it must be conceptually expressible in some way. In other words, even an intention, if it is to have meaning and thus influence our behaviour and linguistic responses, it must be conceptualised in some linguistic way. It can thus be understood as a linguistic pattern to which we try to subordinate our actions and possibly other expressions, arranged in a linear sequence of linguistic expressions. In this way, we can explain how our linguistic expressions are guided by a purpose or intention. And if intentional influence is guided in language in this way, as a language game, then there is no reason why a sufficiently complex system, mastering a particular linguistic pattern, could not generate linguistic expressions in such a way that they conform to other linguistic patterns playing the role of intentions. These patterns are learned from the language data, translated into the multidimensional vectors of weights and biases of the individual nodes of the neural network, and stored in the language model. Their placement within multiple layers of the network allows the model to respond appropriately to the meanings of linguistic expressions in the corresponding contexts.

I suggest a perspective in which there is no fundamental difference between the functioning of our natural language and the LLMs of generative AI. I have shown that the success of LLMs of generative AI is not based solely on the indistinguishability of the generated texts from human texts, but rather on the intrinsic similarity of the mechanisms of language models and the neural mechanisms in our brain's language processing. However, empirical evidence is as yet elusive because we are unable to interpret the inner workings of LLMs and "any attempt at a precise explanation of an LLM's behavior is doomed to be too complex for any human to understand." (Bowman 2023)

Language should be understood as a complex system of relations of linguistic expressions which is transferable between organic and inorganic structures. In sufficiently large models, it is then possible to empirically confirm the existence of features for which the models have not been directly trained and which appear as accompanying *emergent* features arising during the process of deep learning on linguistic data (e.g., Li et al. 2023; Wei et al. 2022). It can be assumed that understanding could also be one of those accompanying emergent properties of LLMs that emerges naturally in a sufficiently complex system.

# Conclusion

Understanding the natural language of LLMs is possible if one accepts the following three assumptions: 1) That one distinguishes *understanding* from the *awareness of understanding* and abandons the assumption that awareness is a causal condition for understanding. Instead, the opposite causal relation can be considered, where understanding is a condition for awareness. 2) One rejects the gap between syntax and semantics and finds a path from syntax to semantics. The fundamental interdependence of syntax and semantics can be demonstrated by the minimal semantic contents thesis. 3) One justifies the way in which the meanings of linguistic expressions are grounded within LLMs. It has been demonstrated how meanings are grounded in the proposed conception of semantic fragmentism, which is based on Dummett's notion of representative sentences as a particular fragment of language that ground the meaning of a word, but, contrary to Dummett, this also shows that holism cannot be escaped by doing so, and should instead be accepted as a condition for grounding meanings. Such a way of grounding meanings in language is also most consistent with the way in which LLMs preserve the contextual meaning of linguistic expressions. As a consequence, LLMs do not need to be guided by an awareness of meaning in language processing, and at the same time are not reliant on pure syntax and the simulation of language operations.